# Hypergraph Laplacian Eigenmaps and Face Recognition Problems

Loc Hoang Tran

Vietnam Aviation Academy

Email: locth@vaa.edu.vn

Abstract: Face recognition is a very important topic in data science and biometric security research areas. It has multiple applications in military, finance, and retail, to name a few. In this paper, the novel hypergraph Laplacian Eigenmaps will be proposed and combine with the *k* nearest-neighbor method and/or with the kernel ridge regression method to solve the face recognition problem. Experimental results illustrate that the accuracy of the combination of the novel hypergraph Laplacian Eigenmaps and one specific classification system is similar to the accuracy of the combination of the old symmetric normalized hypergraph Laplacian Eigenmaps method and one specific classification system.

Keywords: face recognition, hypergraph, Laplacian Eigenmaps, classification

I.   Introduction

Given a relational dataset, the pairwise relationships among objects/entities/samples in this dataset can be represented as the weighted graph. Then, the un-supervised learning techniques such as representational learning methods/dimensional reduction methods and clustering methods and the semi-supervised learning techniques can be applied to this graph. These techniques (the un-supervised learning techniques and the semi-supervised learning techniques) can be formulated as the operations on this graph. The fundamental matrices used in these techniques are the adjacency matrix of the graph and/or the Laplacian matrix of the graph [1,2].

However, assuming the pairwise relationships among the objects/entities/samples in this graph representation is not complete. Let's consider the case that we would like to partition/segment a set of articles into different topics (i.e., clustering problem) [3,4].

Initially, we employ the graph data structure to represent this dataset. The vertices of the graph are the articles. Two articles are connected by an edge (i.e., the relationship) if there is at least one author in common. Finally, we can apply spectral clustering technique [5,6] to this graph to partition/segment the vertices into groups/clusters.

Obviously, we easily see that in this graph data structure, we **ignore** the information whether **one specific author is the author of three or more articles (i.e., the co-occurrence relationship or high order relationship).**

This will lead to **the loss of information**. In other words, this will lead to the low performance (i.e., the low accuracy) of the clustering technique.

To overcome this difficulty, [3,4,10,11,12,13,14,15,16,17,18] try to employ the hypergraph data structure to represent for the above relational dataset. In detail, in this hypergraph data structure, the articles are the vertices, and the authors are the hyper-edges. This hyper-edge can connect more than two vertices (i.e., articles).

**Please note that the simplicial complex is the uniform hypergraph which is one specific type of hypergraph. The uniform hypergraph is the hypergraph where all hyperedges have the same cardinality. However, in this thesis, we mainly discuss the general hypergraph.**

The following figure 1 shows the example of the hypergraph.

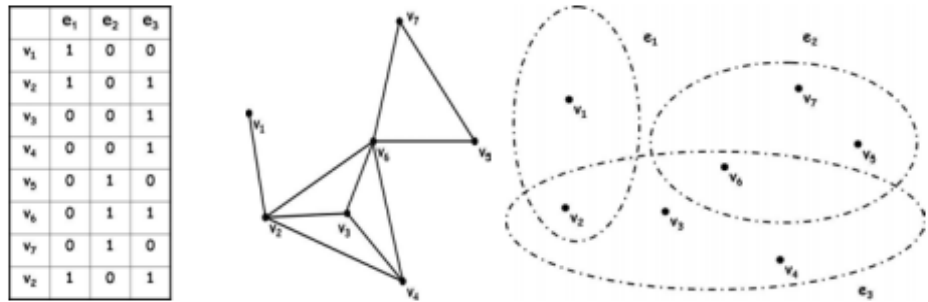

Figure 1. Hypergraph example with 8 vertices and 3 hyper-edges [3].

There are a lot of ways to represent this hypergraph data structure: as the incidence matrix (used in this paper and in [3]) or as the tensor [7,8]. From [3], we recognize that the outcome of the hypergraph-based clustering technique is quite promising. Its performance is better than the performance of the spectral clustering technique (for graph).

To the best of my knowledge, there are five main problems in machine learning/deep learning research field such as:

- Representational learning/Dimensional reduction/…
- Clustering
- Classification
- Link prediction/Recommendation system/…
- (Reinforce Learning)/…

In this paper, we will develop novel methods (i.e., novelty property) which are the novel dimensional reduction methods to solve the classification problem (i.e., the face recognition problem) utilizing the hypergraph data structure.

We will organize the paper as follows: Section 2 will present the preliminary definitions and notations of hypergraph data structure. Section 3 will introduce the un-normalized, random walk, and symmetric normalized hypergraph Laplacian Eigenmaps algorithms in detail. In section 4, we will apply the two proposed hypergraph Laplacian Eigenmaps algorithms (i.e., the combinatorial Laplacian Eigenmaps and symmetric normalized Laplacian Eigenmaps) combined with the *k*-nearest neighbor method and kernel ridge regression method to the face dataset available from [9] and compare their accuracy performance measures. Section 5 will conclude this paper and the future directions of research of these methods will be discussed.

## II. Preliminary definitions and Notations

Given a hypergraph $G = (V, E)$, where $V$ is the set of vertices and $E$ is the set of hyper-edges. Each hyper-edge $e \in E$ is the subset of $V$. Please note that the cardinality of $e$ is greater than or equal to two. In the other words, $|e| \geq 2$, for every $e \in E$. Let $w(e)$ be the weight of the hyper-edge $e$. Then $W$ will be the $R^{|E|*|E|}$ diagonal matrix containing the weights of all hyper-edges in its diagonal entries.

The incidence matrix $H$ of $G$ is a $R^{|V|*|E|}$ matrix that can be defined as follows:

$$h(v, e) = \begin{cases} 1 \text{ if vertex } v \text{ belongs to hyperedge } e \\ 0 \text{ otherwise} \end{cases}$$

From the above definition, we can define the degree of vertex $v$ and the degree of hyper-edge $e$ as follows:

$$d(v) = \sum_{e \in E} w(e) * h(v, e)$$

$$d(e) = \sum_{v \in V} h(v, e)$$

Let $D_v$ and $D_e$ be two diagonal matrices containing the degrees of vertices and the degrees of hyper-edges in their diagonal entries respectively. Please note that $D_v$ is the $R^{|v|*|v|}$ matrix and $D_e$ is the $R^{|e|*|e|}$ matrix.

The un-normalized (combinatorial) hypergraph Laplacian is defined as follows:

$$L = D_v - HWD_e^{-1}H^T$$

The symmetric normalized hypergraph Laplacian (defined in [3,4]) is defined as follows:

$$L_{sym} = I - D_v^{-\frac{1}{2}}HWD_e^{-1}H^TD_v^{-\frac{1}{2}}$$

The random walk hypergraph Laplacian (defined in [3,4]) is defined as follows:

$$L_{rw} = I - D_v^{-1}HWD_e^{-1}H^T$$

## III. Algorithms

Given a set of points $\{x_1, x_2, \ldots, x_n\}$ where $n$ is the total number of points (i.e. vertices) in the hypergraph $G = (V, E)$ and given the incidence matrix $H$ of $G$.

Our objective is to compute the eigenvectors of the three hypergraph Laplacians.

### **Random walk hypergraph Laplacian Eigenmap algorithm**

First, we will give the brief overview of the random walk hypergraph Laplacian Eigenmap algorithm. The outline of this algorithm is as follows:

i. Construct $D_v$ and $D_e$ from the incidence matrix $H$ of $G$
ii. Compute the random walk hypergraph Laplacian $L_{rw} = I - D_v^{-1}HWD_e^{-1}H^T$
iii. Compute all eigenvalues and eigenvectors of $L_{rw}$ and sort all eigenvalues and their corresponding eigenvector in ascending order. Pick the first $k$ eigenvectors $v_2, v_3, \ldots, v_{k+1}$ of $L_{rw}$ in the sorted list. $k$ can be determined in the following two ways:
   a. $k$ is the number of connected components of $L_{rw}$
   b. $k$ is the number such that $\frac{\lambda_{k+2}}{\lambda_{k+1}}$ or $\lambda_{k+2} - \lambda_{k+1}$ is largest for all $1 \leq k \leq n-1$
iv. Let $V \in R^{n*k}$ be the matrix containing the vectors $v_2, v_3, \ldots, v_{k+1}$ as columns and $V$ is the final result

## Un-normalized (combinatorial) hypergraph Laplacian Eigenmap algorithm

Next, we will give the brief overview of the un-normalized (combinatorial) hypergraph Laplacian Eigenmap algorithm. The outline of this algorithm is as follows:

i. Construct $D_v$ and $D_e$ from the incidence matrix $H$ of $G$
ii. Compute the un-normalized hypergraph Laplacian $L = D_v - HWD_e^{-1}H^T$
iii. Compute all eigenvalues and eigenvectors of $L$ and sort all eigenvalues and their corresponding eigenvector in ascending order. Pick the first $k$ eigenvectors $v_2, v_3, \ldots, v_{k+1}$ of $L$ in the sorted list. $k$ can be determined in the following two ways:
   a. $k$ is the number of connected components of $L$
   b. $k$ is the number such that $\frac{\lambda_{k+2}}{\lambda_{k+1}}$ or $\lambda_{k+2} - \lambda_{k+1}$ is largest for all $1 \leq k \leq n-1$
iv. Let $V \in R^{n*k}$ be the matrix containing the vectors $v_2, v_3, \ldots, v_{k+1}$ as columns and $V$ is the final result

## Symmetric normalized hypergraph Laplacian Eigenmap algorithm

Finally, we will give the brief overview of the symmetric normalized hypergraph Laplacian based un-supervised learning algorithm which can be obtained from [3,4]. The outline of this algorithm is as follows:

i. Construct $D_v$ and $D_e$ from the incidence matrix $H$ of $G$
ii. Compute the symmetric normalized hypergraph Laplacian $L_{sym} = I - D_v^{-\frac{1}{2}}HWD_e^{-1}H^TD_v^{-\frac{1}{2}}$
iii. Compute all eigenvalues and eigenvectors of $L_{sym}$ and sort all eigenvalues and their corresponding eigenvector in ascending order. Pick the first $k$ eigenvectors $v_2, v_3, \ldots, v_{k+1}$ of $L_{sym}$ in the sorted list. $k$ can be determined in the following two ways:

a. $k$ is the number of connected components of $L_{sym}$
   b. $k$ is the number such that $\frac{\lambda_{k+2}}{\lambda_{k+1}}$ or $\lambda_{k+2} - \lambda_{k+1}$ is largest for all $1 \leq k \leq n-1$

iv. Let $V \in R^{n*k}$ be the matrix containing the vectors $v_2, v_3, \ldots, v_{k+1}$ as columns and $V$ is the final result

IV. Experiments and Results

In this paper, the set of 120 face samples recorded of 15 different people (8 face samples per people) is the training set. Then another set of 45 face samples of these people is the testing set. This dataset is available from [9]. Then, we will merge all rows of the face sample (i.e. the matrix) sequentially from the first row to the last row into a single big row which is the $R^{1*1024}$ row vector.

Next, the hypergraph Laplacian Eigenmaps algorithms will be applied to faces in the training set and the testing set to reduce the dimensions of the faces. Then the nearest-neighbor method and the kernel ridge regression method will be applied to these new transformed feature vectors.

In this section, we experiment with the above $k$ nearest-neighbor method and kernel ridge regression method in terms of accuracy. The accuracy measure Q is given as follows:

$$Q = \frac{True\ Positive + True\ Negative}{True\ Positive + True\ Negative + False\ Positive + False\ Negative}$$

All experiments were implemented in Python on Google Colab with NVIDIA Tesla K80 GPU and 12 GB RAM. The accuracies of the above proposed methods are given in the following table 1 and table 2.

Table 1: **Accuracies** of the combination of PCA method and the nearest-neighbor method, and the combination of sparse PCA method and the nearest-neighbormethod

|  | Accuracy |
|---|---|
| Combinatorial Laplacian Eigenmaps (d=20) + $k$ nearest neighbor method | 0.67 |
| Combinatorial Laplacian Eigenmaps (d=30) + | 0.60 |
| $k$ nearest neighbor method | |
| Combinatorial Laplacian Eigenmaps (d=40) + $k$ nearest neighbor method | 0.67 |
| Random walk Laplacian Eigenmaps (d=20) + $k$ nearest neighbor method | 0.67 |
| Random walk Laplacian Eigenmaps (d=30) + $k$ nearest neighbor method | 0.60 |
| Random walk Laplacian Eigenmaps (d=40) + $k$ nearest neighbor method | 0.67 |
| Symmetric normalized Laplacian Eigenmaps (d=20) + $k$ nearest neighbor method | 0.67 |
| Symmetric normalized Laplacian Eigenmaps (d=30) + $k$ nearest neighbor method | 0.60 |
| Symmetric normalized Laplacian Eigenmaps (d=40) + $k$ nearest neighbor method | 0.67 |

Table 2: **Accuracies** of the combination of PCA method and the kernel ridge regression method, and the combination of sparse PCA method and the kernel ridge regression method

| | Accuracy |
|---|---|
| Combinatorial Laplacian Eigenmaps (d=20) + kernel ridge regression method | 0.73 |
| Combinatorial Laplacian Eigenmaps (d=30) + kernel ridge regression method | 0.71 |
| Combinatorial Laplacian Eigenmaps (d=40) + kernel ridge regression method | 0.73 |
| Random walk Laplacian Eigenmaps (d=20) + $k$ kernel ridge regression method | 0.73 |
| Random walk Laplacian Eigenmaps (d=30) + $k$ kernel ridge regression method | 0.71 |
| Random walk Laplacian Eigenmaps (d=40) + $k$ | 0.73 |
| kernel ridge regression method | |
| Symmetric normalized Laplacian Eigenmaps (d=20) + kernel ridge regression method | 0.73 |
| Symmetric normalized Laplacian Eigenmaps (d=30) + kernel ridge regression method | 0.71 |
| Symmetric normalized Laplacian Eigenmaps (d=40) + kernel ridge regression method | 0.73 |

From the above tables 1 and 2, we recognize the accuracies of the combination of the two proposed Laplacian Eigenmaps and one specific classification system are like the accuracy of the combination of the symmetric normalized Laplacian Eigenmaps and one specific classification system.

V.  Conclusions

In this paper, two novel hypergraph Laplacian Eigenmaps have been proposed. The accuracy of the combination of the two proposed hypergraph Laplacian Eigenmaps with *k* nearest neighbor method and with kernel ridge regression method are like the accuracy of the combination of the old symmetric normalized hypergraph Laplacian Eigenmaps with *k* nearest neighbor method and the kernel ridge regression method.